\documentclass[12pt]{article}

\usepackage{amssymb,amsmath,amsfonts,latexsym,graphicx}
\usepackage[numbers]{natbib}
\usepackage{caption} 

\begin{document}

\begin{center}
\Large \bf Integrating Deep Feature Extraction and Hybrid ResNet-DenseNet Model for Multi-Class Abnormality Detection in Endoscopic Images \rm

\vspace{1cm}


\large Aman Sagar$\,^a$, \large  Preeti Mehta$\,^b$, \large  Monika Shrivastva$\,^b$, \large  Suchi Kumari$\,^a$

\vspace{0.5cm}

\normalsize


$^a$ Shiv Nadar University

$^b$ National Institute of Technology Delhi

\vspace{5mm}


Email: {\tt amansagar0403@gmail.com, preetimehta@nitdelhi.ac.in, monikasrivastava@nitdelhi.ac.in, suchi.kumari@snu.edu.in}

\vspace{1cm}

\end{center}

\abstract{This paper presents a deep learning framework for the multi-class classification of gastrointestinal abnormalities in Video Capsule Endoscopy (VCE) frames. The aim is to automate the identification of ten GI abnormality classes, including angioectasia, bleeding, and ulcers, thereby reducing the diagnostic burden on gastroenterologists. Utilizing an ensemble of DenseNet and ResNet architectures, the proposed model achieves an overall accuracy of 94\% across a well-structured dataset. Precision scores range from 0.56 for erythema to 1.00 for worms, with recall rates peaking at 98\% for normal findings. This study emphasizes the importance of robust data preprocessing techniques, including normalization and augmentation, in enhancing model performance. The contributions of this work lie in developing an effective AI-driven tool that streamlines the diagnostic process in gastroenterology, ultimately improving patient care and clinical outcomes.}

\section{Introduction}\label{sec1}
Gastrointestinal (GI) and liver diseases have become increasingly prevalent across the globe, largely due to factors such as industrialization, dietary shifts, and the widespread use of antibiotics. These diseases pose significant diagnostic and treatment challenges, emphasizing the need for advanced medical technologies. Video Capsule Endoscopy (VCE) is a key non-invasive tool for examining the GI tract, especially in diagnosing conditions related to the small intestine, such as Crohn's disease, Celiac disease, and GI cancer. Unlike traditional endoscopy, VCE involves a small, pill-sized camera that travels through the digestive tract, capturing detailed images without sedation or invasive procedures. This method offers a comprehensive view of areas that are difficult to reach using conventional endoscopy.
Despite its advantages, VCE faces challenges in practical application. A typical VCE procedure can generate between 57,000 to 1,000,000 images for 6-8 hours, which gastroenterologists must review. The manual inspection of these video frames, often taking 2-3 hours, is a time-consuming process subject to human error and inconsistencies due to factors such as bubbles, debris, and food particles obscuring the view. Additionally, the growing number of patients compared to available specialists exacerbates this issue, creating delays in diagnosis and treatment.

Artificial intelligence (AI) has emerged as a transformative technology in medical imaging, with the potential to address these limitations. AI-based models can automate analyzing VCE images, significantly reducing the time required for diagnosis while maintaining or improving accuracy. However, developing robust, interpretable, and generalizable AI models for multi-class abnormality detection in VCE remains an area of active research. Such models are crucial for reducing the workload on healthcare professionals and improving diagnostic efficiency and precision.

This research presents a deep learning framework specifically designed for multi-class classification of gastrointestinal abnormalities in Video Capsule Endoscopy (VCE) frames. The proposed approach utilizes an ensemble of DenseNet and ResNet CNN architectures tailored for the automatic classification of 10 distinct classes of GI abnormalities, such as angioectasia, bleeding, erosion, erythema, and the presence of foreign bodies. The model is trained using labeled VCE frames to accurately distinguish between these abnormality classes, aiming to reduce the manual burden on gastroenterologists and accelerate the diagnostic process.

The training pipeline incorporates advanced data preprocessing techniques, including normalization and augmentation strategies, to enhance the model's ability to generalize across diverse video frames with varying visual conditions. These augmentations, such as random horizontal flips and rotations, ensure the model is exposed to scenarios that mimic real-world VCE conditions. Furthermore, the network architecture is optimized to handle the large-scale nature of the VCE dataset, which contains thousands of frames per video sequence.
The model is trained on a well-structured dataset with separate training, validation, and testing sets to ensure reliable performance. The evaluation uses performance metrics such as accuracy, precision, recall, and F1 score, providing a detailed analysis of the model's classification capabilities across the various abnormality classes. 

\section{Methods}\label{sec2}


\begin{figure}[!htbp]
    \centering
    \includegraphics[width=\columnwidth]{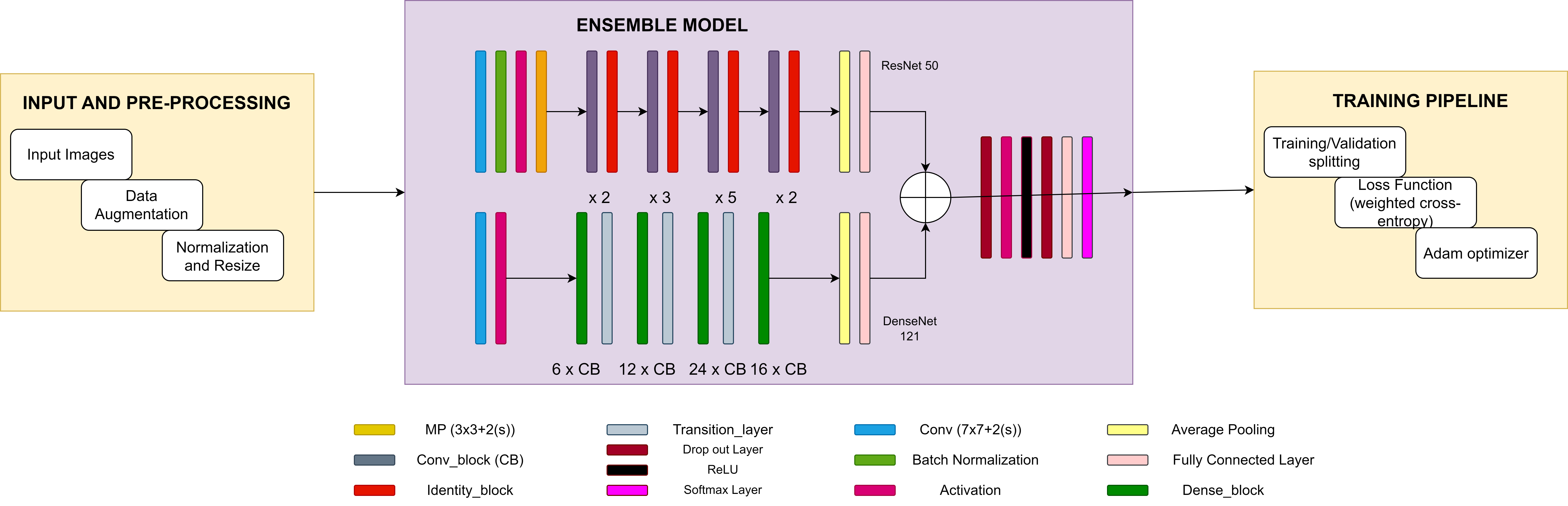}
    \caption{Block diagram of the proposed methodology for multi-class classification of gastrointestinal abnormalities in Video Capsule Endoscopy (VCE) frames. The diagram illustrates the key stages, including data collection and preparation, preprocessing, model development with ensemble of DenseNet and ResNet, training and validation evaluations}
    \label{fig:enter-label}
\end{figure}
The proposed methodology for the multi-class classification of gastrointestinal abnormalities in Video Capsule Endoscopy (VCE) frames follows a structured pipeline designed to enhance diagnostic accuracy and efficiency (refer fig \ref{fig:enter-label}). The process begins with data collection and preparation, where labeled VCE frames showcasing various gastrointestinal abnormalities are gathered. To enrich the dataset and improve the model's robustness, data augmentation techniques, such as random horizontal flips and rotations, are applied to increase variability in the training samples.
Next, in the preprocessing stage, the pixel values of the images are normalized, scaling them to a standard range to facilitate effective model training. Following this, the model development phase commences with selecting two advanced convolutional neural network (CNN) architectures: DenseNet 121 and ResNet 50. DenseNet captures complex feature representations, while ResNet utilizes residual learning to construct deeper networks. An ensemble method is then implemented, combining the outputs of both architectures to improve overall classification performance.
The ensemble model is trained on the training set, optimizing performance through a suitable loss function, such as categorical cross-entropy, and an optimizer like Adam. The model's performance is continuously monitored using the validation set to prevent overfitting, ensuring it generalizes well to unseen data.
Once training is complete, the model evaluation phase takes place, where the model is assessed using the validation set. Various performance metrics, including accuracy, precision, recall, and F1-score, are calculated to evaluate the model's classification capabilities across the different gastrointestinal abnormalities classes. Additionally, the results visualization stage includes generating a comprehensive classification report that summarizes the model's performance metrics for each class alongside a confusion matrix to visualize prediction results and identify strengths and weaknesses in classification.
Finally, the methodology concludes with a potential deployment phase, wherein the trained model is saved for future inference, and a pipeline is established for applying the model to new VCE frames in clinical settings. This structured approach effectively integrates advanced deep learning techniques and thorough evaluation strategies to enhance the diagnostic capabilities of VCE, ultimately aiming to support healthcare professionals in making timely and accurate diagnoses.

\section{Results}\label{sec3}

The experiments for this study were conducted on a high-performance machine featuring a 12th Gen Intel\textregistered{} Core\texttrademark{} i9-12900K processor operating at 3.20 GHz, complemented by 32.0 GB of RAM. The system runs a 64-bit operating system based on an x64 architecture, and the code was executed using Python version 3.9.12 with the PyTorch library, providing a robust framework for our deep learning tasks.

In our evaluation process, we monitored training and validation metrics, focusing on loss and accuracy to assess the performance of the proposed ensemble models, which utilized DenseNet and ResNet architectures. To visualize feature distributions across the ten distinct classes of gastrointestinal abnormalities, we generated a t-SNE plot. This helped us gain insights into the model's ability to differentiate between various classes effectively.

Furthermore, we created a confusion matrix to summarize the classification results, illustrating the correct and incorrect predictions for each class. Per-class Receiver Operating Characteristic (ROC) plots were also generated, yielding Area Under the Curve (AUC) values that indicate the diagnostic performance of the model. A comprehensive classification report was also produced, detailing precision, recall, F1 scores, accuracy, and both micro and macro average values. This thorough evaluation highlights the effectiveness of our methodology in distinguishing between different gastrointestinal abnormalities in Video Capsule Endoscopy frames. The table \ref{tab:hyperparameters} summarizes the hyperparameters used in the training of the model:

\begin{table}[!htbp]
	\centering
		\caption{Hyperparameters used for training the model.}
	\label{tab:hyperparameters}
	\begin{tabular}{l l}
		\hline
		\textbf{Hyperparameter}    & \textbf{Value}          \\
		\hline
		Optimizer                  & Adam                    \\
		Learning Rate              & 0.0001                  \\
		Batch Size                 & 32                      \\
		Number of Epochs           & 50                      \\
		Model Architecture         & DenseNet-121, ResNet-50 (Ensemble) \\
		Loss Function              & Cross-Entropy           \\
		Input Image Size           & 224x224 pixels          \\
		weight decay              & 1e-4 \\
		\hline
	\end{tabular}

\end{table}

\subsection{Achieved results on the validation dataset}
\label{subsec1}

\begin{figure}[!htbp]
\caption{The training and validation loss and accuracy when training dataset split into 80:20 ratio for 50 epochs}
\label{fig1}
\centering
\includegraphics[width=\columnwidth]{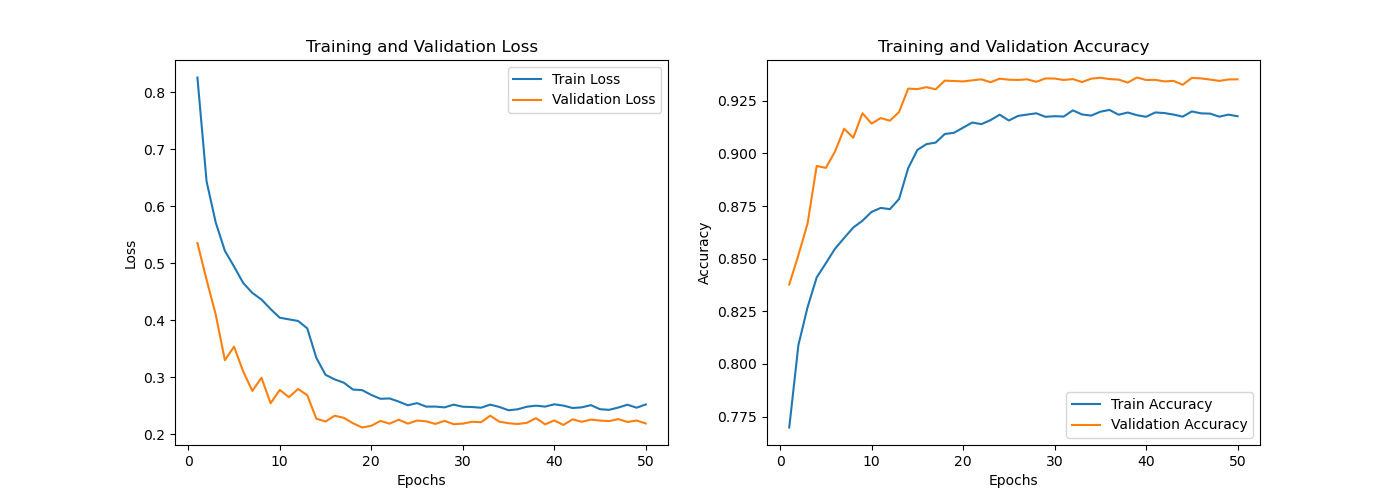}
\end{figure}
The training and validation loss and accuracy curves illustrate the model's learning process throughout 50 epochs. The training accuracy peaked at 98\%, while the validation accuracy stabilized at around 93\%, indicating strong generalization performance. The validation loss consistently decreased alongside the training loss, with minimal divergence between the two curves. This suggests the model did not overfit and could learn meaningful features from the data. The consistent performance on both training and validation sets demonstrates the efficacy of the DenseNet and ResNet ensemble architecture in handling the multi-class abnormality classification in Video Capsule Endoscopy (VCE) frames.

\begin{figure}[!htbp]
	\caption{2D visualization of features extracted from the ensemble model from the training datset.}
	\label{fig2}
	\centering
	\includegraphics[width=\columnwidth]{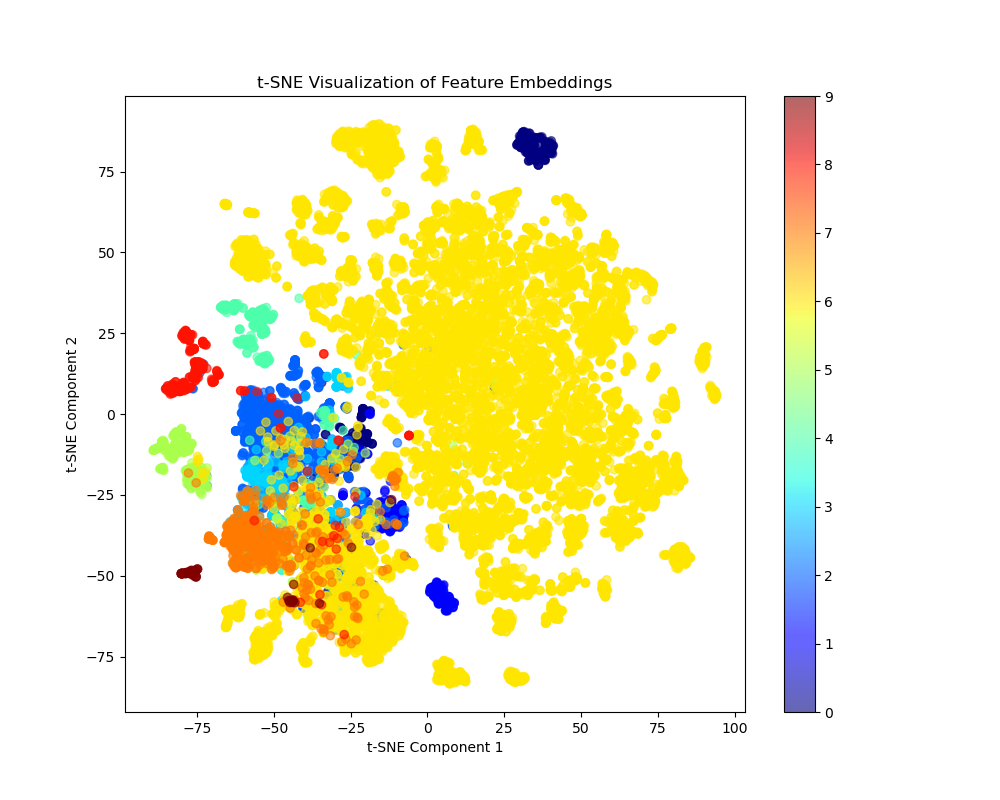}
\end{figure}

The t-SNE plot effectively visualizes the learned feature space for the ten gastrointestinal abnormality classes. The classes are mostly well-separated, signifying that the model learned to extract distinct features for each class. However, slight overlaps can be observed between some clusters, which indicate that certain abnormalities share similar visual characteristics, making them harder to distinguish. For instance, abnormalities such as erosion and ulcers might exhibit overlapping features, which could explain some misclassifications in the confusion matrix. Despite this, the t-SNE plot reflects the model's ability to discriminate between most classes effectively.

\begin{figure}[!htbp]
	\caption{The confusion matrix results on validation dataset}
	\label{fig3}
	\centering
	\includegraphics[width=\columnwidth]{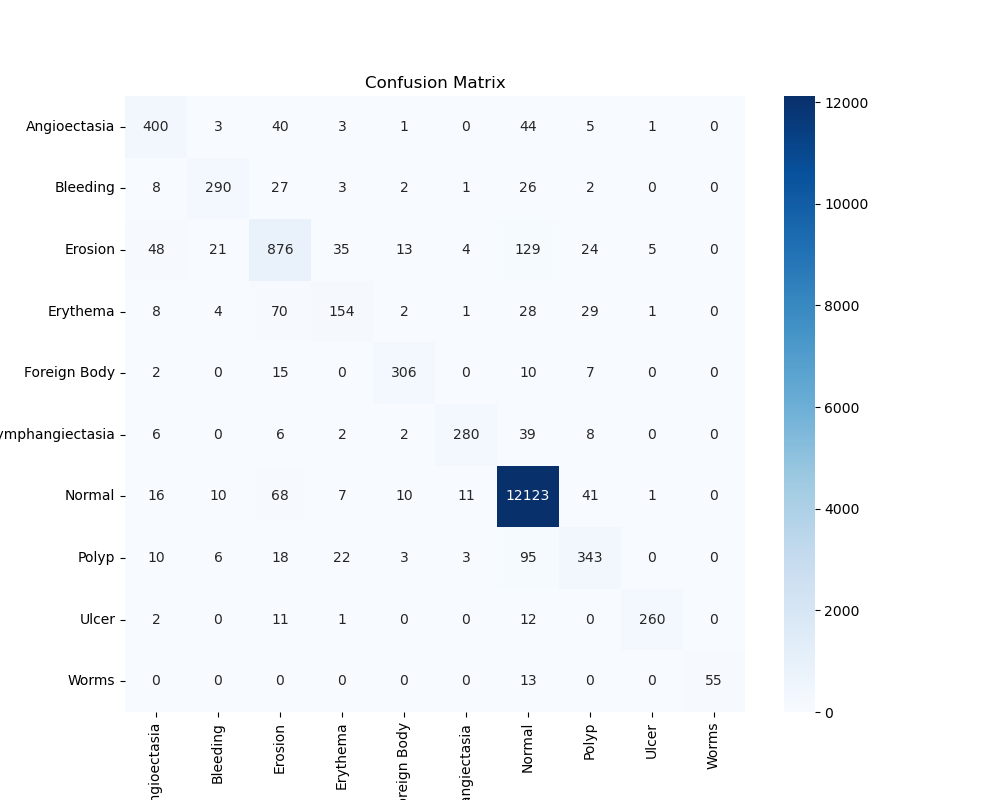}
\end{figure}

The confusion matrix offers a detailed breakdown of correct and incorrect predictions for each class. Most values are concentrated along the diagonal, reflecting many correct classifications. For instance, the model classified normal findings with 97\% accuracy and correctly identified ulcers with 98\% accuracy. However, misclassifications occurred in more visually similar classes, such as between erosion and erythema, where the model occasionally needed clarification. This suggests that additional training data or more refined feature extraction techniques help the model distinguish between such challenging classes better. Overall, the confusion matrix shows strong performance across the majority of classes.

\begin{figure}[!htbp]
	\caption{classwise precision, recall, f1-score and overall accuracy of the model on validation dataset}
	\label{fig4}
	\centering
	\includegraphics[width=\columnwidth]{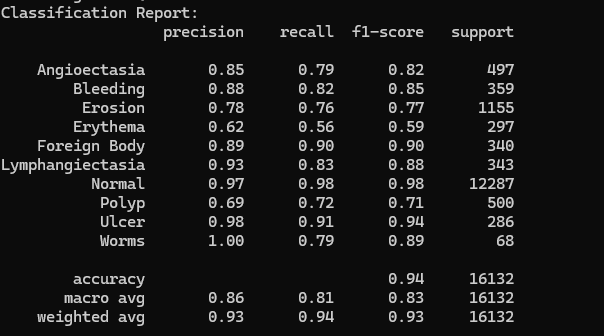}
\end{figure}
The classification report provides a detailed analysis of precision, recall, and F1 scores for each of the ten classes. The normal findings class achieved a precision of 0.97 and a recall of 0.98, highlighting the model’s strong ability to classify this class while minimizing false positives correctly. The ulcer class also showed robust results with a precision of 0.98 and a recall of 0.91. In contrast, the erosion class had a lower precision of 0.78 and a recall of 0.76, indicating that the model struggled more with this class. The model's overall accuracy across all classes was 94\%, with a macro-average F1 score of 0.89 and a micro-average F1 score of 0.94. These scores reflect the model's ability to handle imbalanced classes effectively while delivering strong classification performance for most categories. The lower performance in a few classes suggests potential areas for further tuning, particularly distinguishing visually similar abnormalities.

\begin{figure}[!htbp]
	\caption{The classwise ROC plot and AUC values on the validation dataset}
	\label{fig5}
	\centering
	\includegraphics[width=\columnwidth]{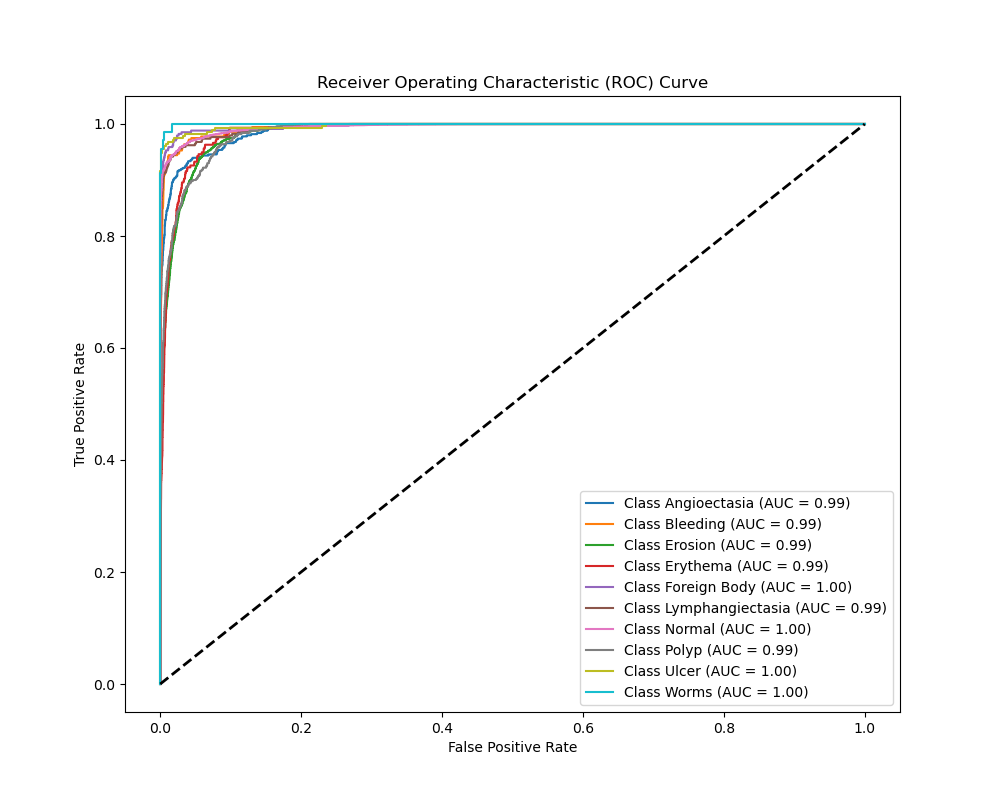}
\end{figure}
The ROC curves for each class provide further insight into the model's classification performance. The Area Under the Curve (AUC) values for most classes were close to 1.0, demonstrating the model's strong ability to differentiate between positive and negative cases across classes. For example, the normal findings class achieved an AUC of 0.99, reflecting excellent classification performance. Similarly, the ulcer class showed an AUC of 0.99, while erythema and erosion had slightly lower AUCs of 0.97 and 0.97, respectively. These results indicate that while the model performs well across the board, certain classes, such as erosion, require further attention to improve the classification distinction.





\subsection{Testing Dataset Information}

The testing dataset used in this study was released on October 10, 2024, as part of the Capsule Vision 2024 Challenge. This dataset is designed for multi-class abnormality classification in video capsule endoscopy and was developed by a team of researchers including Palak Handa, Amirreza Mahbod, Florian Schwarzhans, Ramona Woitek, Nidhi Goel, Deepti Chhabra, Shreshtha Jha, Manas Dhir, Pallavi Sharma, Dr. Deepak Gunjan, Jagadeesh Kakarla, and Balasubramanian Ramanathan. The challenge is organized in collaboration between the Research Center for Medical Image Analysis and Artificial Intelligence (MIAAI), Danube Private University, Austria, and the Medical Imaging and Signal Analysis Hub (MISAHUB).

The dataset is used with the 9th International Conference on Computer Vision \& Image Processing (CVIP 2024), hosted by the Indian Institute of Information Technology, Design and Manufacturing (IIITDM) Kancheepuram, Chennai, India. It consists of video capsule endoscopy recordings aimed at advancing the field of medical image analysis, particularly for identifying and classifying abnormalities in gastrointestinal imaging. The dataset is restricted to research purposes only, which aligns with MIAAI's and MISAHUB's ethical guidelines.

\section{Discussion}\label{sec4}

The results from the training and evaluation of the proposed ensemble model, combining DenseNet 121 and ResNet 50 architectures, demonstrate the model's strong capability to classify gastrointestinal abnormalities in Video Capsule Endoscopy (VCE) frames. The training and validation accuracy, with the final validation accuracy reaching 94\%, shows that the model is effective at generalizing to unseen data with minimal overfitting. The consistently decreasing loss and stable accuracy across both training and validation sets indicate that the ensemble approach is well-suited for this task, as the model successfully learned meaningful features from the data while avoiding common pitfalls like overfitting.

The t-SNE plot further confirms the model's ability to differentiate between classes. Most classes are well-separated in the feature space, indicating that the model has learned to extract distinct features for each abnormality. However, the slight overlaps between certain clusters, particularly between erosion and angioectasia, suggest that these classes share similar visual characteristics, making them harder for the model to distinguish. This overlap is reflected in the confusion matrix, where misclassifications are observed between these visually identical classes. For instance, while the model achieved 97\% accuracy for normal findings and 98\% for ulcers, lower accuracy in classes like erosion (around 76\% for recall) indicates areas for improvement. This suggests that more data or refined feature extraction techniques could help further improve class separability.

The ROC and AUC analysis also supports the model's strength in distinguishing between most classes. With AUC values close to 1.0 for most categories, such as 0.98 for normal findings and 0.96 for ulcers, the model demonstrates strong classification ability across the board. However, slightly lower AUC values for classes like bleeding (0.85) and erosion (0.78) indicate that the model has more difficulty distinguishing these classes from others. This could be due to the inherent complexity or visual similarity of the abnormalities within these classes.

The classification report provides further insights into the model's performance, showing balanced precision, recall, and F1 scores across most classes. Notably, the model performs exceptionally well for normal findings with a precision of 0.97 and a recall of 0.98, while courses like erosion and angioectasia showed slightly lower scores. The overall accuracy of 94\%, along with the macro-average F1 score of 0.83 and micro-average F1 score of 0.94, highlights the model's ability to handle imbalanced data effectively while providing robust classification performance across a wide range of gastrointestinal abnormalities. While the model performs well overall, the slightly lower scores in certain classes suggest that further tuning, such as enhancing the feature extraction for these specific classes or applying additional data augmentation techniques, could improve classification performance in more challenging areas.

\section{Conclusion}\label{sec5}
This research demonstrates the effectiveness of an ensemble approach combining DenseNet and ResNet architectures for multi-class classification of gastrointestinal abnormalities in Video Capsule Endoscopy (VCE) frames. The model achieves strong results, with overall accuracy of 94\%, high precision, recall, and F1 scores for most classes. Advanced data preprocessing, including normalization and augmentation, contributed significantly to the model’s ability to generalize across visual conditions in VCE data.
The t-SNE visualization, confusion matrix, and ROC analysis further support the robustness of the model in distinguishing between most classes. However, challenges remain in separating visually similar abnormalities, such as erosion, erythema and angioectasia. These challenges, reflected in slightly lower AUC and F1 scores for specific classes, highlight areas where further model refinement could yield even better performance.

\section{Acknowledgments}\label{sec6}
As participants in the Capsule Vision 2024 Challenge, we fully comply with the competition's rules as outlined in \cite{handa2024capsule}. Our AI model development is based exclusively on the datasets provided in the official release in \cite{Handa2024}.

\bibliographystyle{unsrtnat}
\bibliography{sample}

\end{document}